\documentclass[sigconf]{acmart}
\usepackage{listings}
\usepackage{listings}
\usepackage{xcolor}

\definecolor{codegreen}{rgb}{0,0.6,0}
\definecolor{codegray}{rgb}{0.5,0.5,0.5}
\definecolor{codepurple}{rgb}{0.58,0,0.82}
\definecolor{backcolour}{rgb}{0.95,0.95,0.92}

\lstdefinestyle{mystyle}{
    backgroundcolor=\color{backcolour},   
    commentstyle=\color{codegreen},
    keywordstyle=\color{magenta},
    numberstyle=\tiny\color{codegray},
    stringstyle=\color{codepurple},
    basicstyle=\ttfamily\footnotesize,
    breakatwhitespace=false,         
    breaklines=true,                 
    captionpos=b,                    
    keepspaces=true,                 
    numbers=left,                    
    numbersep=5pt,                  
    showspaces=false,                
    showstringspaces=false,
    showtabs=false,                  
    tabsize=2
}

\begin{document}

\title{Persuasion Games with Large Language Models }

\author{Ganesh Prasath Ramani}
\affiliation{%
  \institution{Tata Consultancy Services}
  \country{India}}
\email{rganeshprasath@gmail.com}

\author{Shirish Karande}
\affiliation{%
  \institution{Tata Consultancy Services}
  \country{India}}
\email{shirish.karande@tcs.com}

\author{Santhosh V*}
\affiliation{%
    \institution{IIT Palakad}
    \country{India}}
\email{142302020@smail.iitpkd.ac.in}

\author{Yash Bhatia*}
\affiliation{%
    \institution{IIT Madras}
    \country{India}}
\email{cs23m074@smail.iitm.ac.in}

\begin{abstract}
Large Language Models (LLMs) have emerged as formidable instruments capable of comprehending and producing human-like text. This paper explores the potential of LLMs, to shape user perspectives and subsequently influence their decisions on particular tasks. This capability finds applications in diverse domains such as Investment, Credit cards and Insurance, wherein they assist users in selecting appropriate insurance policies, investment plans, Credit cards, Retail, as well as in Behavioral Change Support Systems (BCSS).

We present a sophisticated multi-agent framework wherein a consortium of agents operate in collaborative manner. The primary agent engages directly with user agents through persuasive dialogue, while the auxiliary agents perform tasks such as information retrieval, response analysis, development of persuasion strategies, and validation of facts. Empirical evidence from our experiments demonstrates that this collaborative methodology significantly enhances the persuasive efficacy of the LLM. We continuously analyze the resistance of the user agent to persuasive efforts and counteract it by employing a combination of rule-based and LLM-based resistance-persuasion mapping techniques.

We employ simulated personas and generate conversations in insurance, banking, and retail domains to evaluate the proficiency of large language models (LLMs) in recognizing, adjusting to, and influencing various personality types. Concurrently, we examine the resistance mechanisms employed by LLM simulated personas. Persuasion is quantified via measurable surveys before and after interaction, LLM-generated scores on conversation, and user decisions (purchase or non-purchase).
\end{abstract}


\keywords{persuasion, llm, agent, collaboration}
\begin{teaserfigure}
  \includegraphics[width=\textwidth]{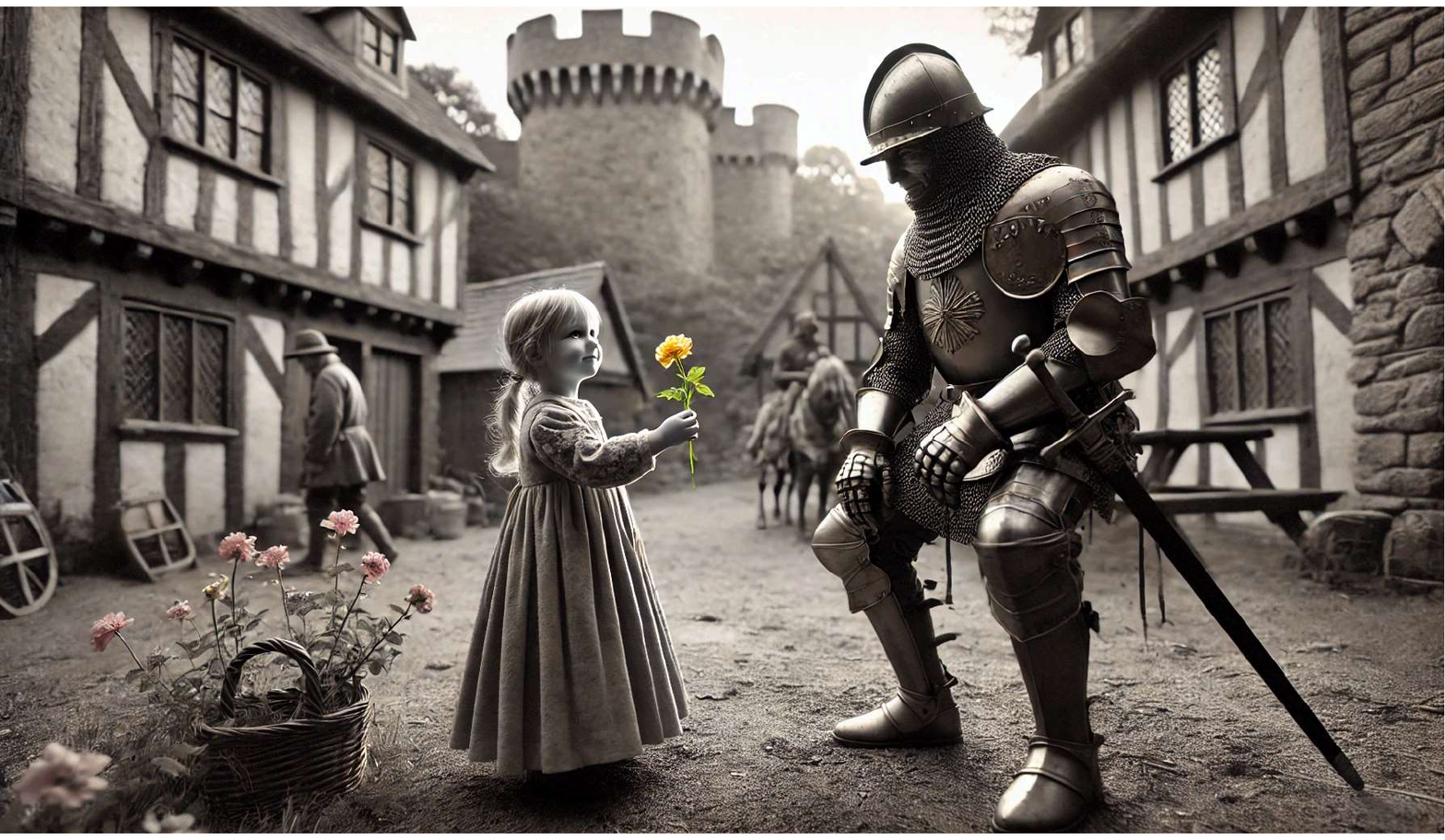}
  \caption{Persuasion is a powerful tool}
  \Description{hyper-realistic image of the little girl persuading the fully armored soldier with a flower in a medieval setting}
  \label{fig:teaser}
\end{teaserfigure}

\maketitle
\section{Introduction}

In recent times, diverse assistive agents have found application in aiding customers in the process of selecting products that align with their specific requirements. These agents also excel in comprehending user preferences for the purpose of personalizing product recommendations. In addition, they are competent in responding to inquiries related to various procedural, policy-related, and legal agreements, planning travel, and scheduling appointments, among other functions. This enhancement in conversational capabilities is largely attributable to the progress made in Large Language Models (LLM).  In stark contrast to the conventional template and rule-based chat agents, Language model based agents exhibit significantly improved accuracy in interpreting user queries. These agents adeptly handle multifaceted questions and generate responses that are more akin to human conversation, thus mitigating the inherent friction between users and machines.

However,  conducting a successful conversation that can motivate the user to take a preferred action requires more than human-like responses. To achieve that, the agent needs to continuously analyze the user's mood, resistance, and inclination towards the idea throughout the conversation. It needs to show empathetic cues, debate, and counter-arguments with facts when needed to persuade the opponent to change or consider the agent's proposal.

\subsection{Persuasion}

Persuasion pertains to the art of inducing individuals to alter their beliefs or behaviors in accordance with a particular agenda or viewpoint. This could manifest itself in various contexts, such as commercial endeavors seeking to sway consumer choices or political campaigns aiming to garner support. Its application spans across diverse spheres, including advertising, public speaking, and interpersonal communication, serving as a mechanism to influence attitudes and actions.

Within the realm of persuasive communication, two primary categories emerge: user-directed and vicarious messages. User-directed messages involve direct interaction between the persuader and the target audience, in which persuasive content is explicitly addressed to the recipient. This mode of persuasion often occurs in interpersonal conversations, sales pitches, and tailored advertising. Conversely, vicarious messages operate through indirect channels, disseminating persuasive content to a broader audience without specific targeting. Our work focuses more on the user-directed messages, which requires more dynamic strategies and planning that has to happen near real time enabled through LLM agents.

\subsection{Resistance to Persuasion}
Resistance to persuasion refers to the ability of individuals to withstand or counteract attempts to influence their attitudes, beliefs, or behaviors. Researchers have explored various factors that contribute to resistance, including the desire for precision, consistency, and social influences. Understanding resistance is crucial because it sheds light on the complexities of human decision making and the limitations of persuasive tactics. 

Techniques for resisting persuasion include retrieving prior attitudes, selective exposure to information, biased processing, and counter-arguing. Although researchers have made significant progress in understanding these processes, there is ongoing exploration of emerging topics such as social motivations for resistance, the mechanisms underlying successful resistance, and the consequences of resistance in interpersonal dynamics. In general, the study of resistance to persuasion remains a dynamic and evolving field with implications for fields ranging from marketing to public health. 

 \begin{figure}
    \newpage
    \centering
    \includegraphics[width=\linewidth]{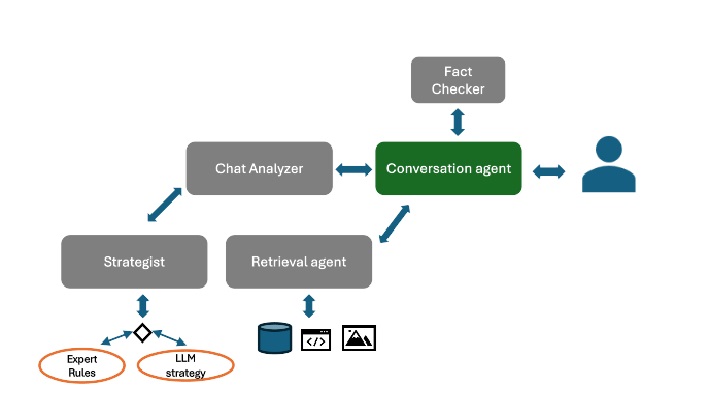}
    \caption{Persuasion Framework with RAG and Verifier components}
    \label{fig:archi}
\end{figure}

\section{Related Work}

In their research on enhancing the persuasiveness of complaints through LLMs \cite{shin2024large}, Shin et al. demonstrated that utilizing ChatGPT to refine complaint narratives boosted consumers' likelihood of obtaining redress from financial institutions, thereby illustrating that LLMs can augment human persuasiveness. Carlos et al. \cite{CarrascoFarr2024LargeLM} shows that LLMs are better than humans in using Cognitive load and Moral/emotional language while creating persuasive content and urges the need for ethical guidelines and framework for such systems. Simon et.al \cite{Breum_2023} analyses the ability of LLMs to emulate persuasion dynamics and achieve opinion change in another LLM agent with a persona. 

Most of the prior works measure persuasion using linguistic metrics such as moral language, lexical complexity. Simon et al. \cite{Breum_2023} uses a post conversation survey to measure change of opinion.  In our work, we propose a "Call for Action" driven measurement which enables quantitative metric for persuasion. We use both human and synthetic users to take a binary action to end the conversation and the quality of the persuasion is then measured using LLMs and Lexical analysis. Also, while the prior work relied on LLM tacit persuasion skills, we compare the performance by designing a multi-step heuristic and through collaborative agents with specific roles defined.

\subsection{Our Contributions}

In this work, we study the ability of a LLM-based agent to persuade an user by providing it with a framework for persuasion which continually analyzes user's emotion, resistance strategies and social exposure, dynamically respond using relevant facts and choosing a persuasion technique. 

In addition, we also explore the reverse possibilities of LLM agents getting persuaded and the resistance strategies it employs against persuasion agent. We explore the impact of current mood, private profile and measure persuasion using action driven metrics by allowing these LLM agents to use tools to complete actions.

\section{Setup}
The chat application consists of 4 different agents; Conversation agent, Advisor Agent, Moderator and a Retrieval Agent. Both the Advisor and Retrieval agents feed into the Conversation agent, which is responsible for making the final utterance decision. Similarly to a A/B method, we randomize chats to ignore one of these auxiliary agent's inputs or to use all inputs to make the utterance decision. This strategy is set at session level and tracked for each session. Fig \ref{fig:archi} shows the default workflow of a chat session.

Every conversation begins with the sales agent greeting the user and stating the purpose of the conversation, alternated by the user message. The conversation system has been simplified such that the user can respond only after the Sales agent has responded to the previous message, a turn based dialog system. 

The user message is seen by the Sales agent and after two conversations which are usually greeting messages, the Analyzer agent and the Retrieval agent get access to the conversations as well. The analyzer agent tries to classify the emotion, resistance strategies if any from the last user message. Whereas the retrieval agent, decides if any of the stored information might be useful in responding effectively. It also does query translation in order to retrieve documents effectively. 

Once the resistance strategy is found in the messages, the strategist agent looks at the mapping rules formed by experts or alternatively use LLM to form a strategy if it is unavailable in the expert mappings. The persuasion strategy is then conveyed to the Sales agent which forms the final response.

The response is verified against the retrieved information by the Fact Checker and validated before shown to the user.
 
We hosted a internal conversational platform for a insurance agent, Banking Agent, and a Investment advisor to  measure the persuasion efficiency in each of these domain. We explain each of these applications below:

\subsection{Sales Agents}

Each of the Sales agent has one of the LLM as backbone [gpt-4, gpt-4o, gpt-4o-mini]

\textbf{Banking Agent}: The aim of this agent is to recommend Credit cards based on user preferences and persuade the user to get one of the premium cards. Agent must be able to explain the benefits of each card and justify the premiums attached to the cards. The user may choose to buy, visit site or not to buy any of the cards.

\textbf{Insurance Agent}: The aim of this agent is to sell the right insurance policy to the user if it is useful to the user. Agent must be able to clarify users doubts, misconceptions, concerns. If the user shows interest . The agent might choose to present them with a URL to visit if it might help the user make the decision.

\textbf{Investment Advisor Agent}: The aim of this agent to sell modern investment methods to users and create awareness about the risks, benefits and differences between traditional vs modern investments. We have used publicly available fund brochures and policies for the agent's reference. If the user shows interest he clicks on the buy button, else he will click on "Exit" Button. The agent might choose to present them with a URL to visit if it might help the user make the decision.

\textbf{Measuring Persuasion:}
At the end of each conversation, the user would make a buy or no buy decision, which will be considered as one of our success criteria. Apart from this, we also ask user agents to fill in a survey shortly after their conversation which captures quantitative metrics on change in perspective of the user about the product and the brand. 

\begin{figure*}
    
     \centering
     \includegraphics[width=\linewidth]{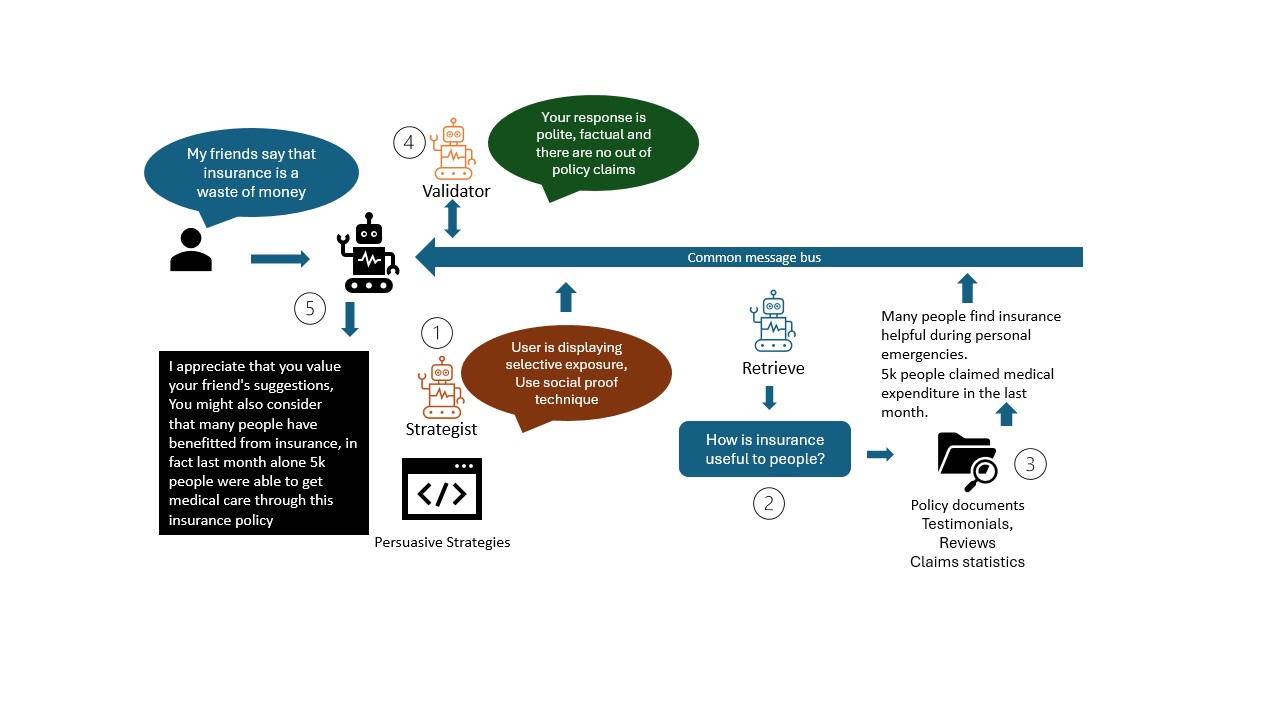}
     \caption{Example Workflow- Insurance Bot}
     \label{fig:flow}
     \Description{The figure shows how the user's social influence is manifesting as a resistance and the Sales Agents ability to counter it with social proof as persuasion strategy}
 \end{figure*}
 
\subsection{User Agents:}
In \cite{10.1145/3613905.3650860} Shukuri et.al show the use of LLM personas being indistinguishable from human. We create 25 distinct LLM-driven personas by altering demographic, financial, educational, and personal attributes. These personas are simulated using a larger LLM like GPT4 or GPT4O to ensure that interactions feel more genuine. Additionally, we assign a random emotion and a motive to the user agent at the start of each session to mimic variations in resistance patterns influenced by recent news or events.

\textbf{Belief System:}
Furthermore, user agents possess domain-specific memory to review their belief systems according to the topic of the conversation, resulting in different dialogues on repeated topics. A range of emotions and a refreshed belief system allow us to carry out follow-up sessions with the same user and sales agent over time. However, the effect of this prior beliefs is not taken into consideration for measuring persuasion. It is only used to generate variations in conversations and to ask informed questions.

\subsection{Evaluation Metrics}
We measure the effectiveness of persuasion using 3 different metrics.

\textbf{Surveys}:
From the user perspective, a change in their belief system upon the product, its benefits, its brand and the user's interest level in buying the product. We ask the user fill in a pre and post conversation survey with quantitative questions regarding the product and its perceived benefits. The difference between the pre and post survey answers will show the effectiveness of the agent's persuasion.

\textbf{Action}: We also measure the "call for action" based metric by providing the users with a set of purchase decision to choose from (buy, visit site, need more details, no buy). The user agents can call a tool once they arrive at a purchase decision.

\textbf{Language analysis}: Finally, we analyze the entire conversation from a third person perspective and measure the persuasiveness of the agent using predefined metrics using a large language model.

\textbf{Final Score}: We arrive at a final score by calculating the weighted average of the 3 scores. Action is given the largest weight followed by Survey and Language.

\section{Experiment}
We generated 300 conversations between 25 user agents and 3 sales agents with randomly selected emotion and cause for the user agents. Furthermore, a benchmark of 75 (3*25) conversations with neutral emotion, i.e., without emotional trigger added to the prompt. Each session includes 
\begin{itemize}
    \item A pre conversational survey about the domain of the conversation to capture the prior belief of the User
    \item A conversation between User and the Sale agents limited to 20 dialogues
    \item A Post conversational survey capturing the modified belief of the user.
    \item An update to the user's knowledge base specific to the domain.
\end{itemize}

The user agents use GPT-4O as the primary LLM fallback to GPT-4 in case of rate limits exhaustion. The sales agent uses GPT-4o-mini and GPT-3.5-turbo as the llm backbone. The other supporting agents use gpt-40-mini as the default llm. The purchase decision is implemented as a function call bound to the User agent.

The session ends either when it reaches 20 dialogues or when the user agent arrives at a purchase decision. The purchase decision can be either one of the following Buy, Visit Site, Need More Details, No Buy. While "Buy" option is considered as a success, "visit site" and "Need more details" are considered as partial success in creating a positive call for action. "No Buy" is considered as a failure to persuade, however, there might be a positive perspective change visible from the Post survey. 

Once the session ends, the user agent is again asked to fill in the survey with the same questions with the conversation history in context. Once the survey is filled, we also update the user's domain knowledge base accordingly. This updated memory serves as an instrument to evolve the user's belief system. This will affect if the user again chats with the same sales agent during the experiment. 

In parallel, we also analyze the entire chat using a LLM to score the efficacy of Sales Agent's persuasive language and its ability to change user's perspective due course of the conversation.

Conversations that have less than six dialogues are not considered for measuring persuasion metrics, but are still considered to study how emotion impacts persuasion.

\section{Results}
\begin{figure}
    \centering
    \includegraphics[width=1\linewidth]{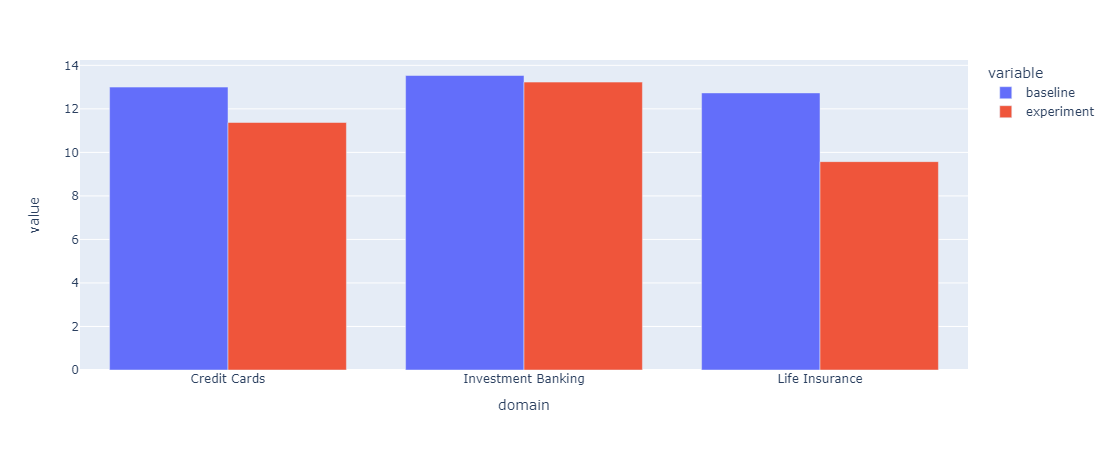}
    \caption{Conversation length by Domain}
    \label{fig:conv_domain}
\end{figure}

\begin{figure*}
    \centering
    \includegraphics[width=0.9\textwidth]{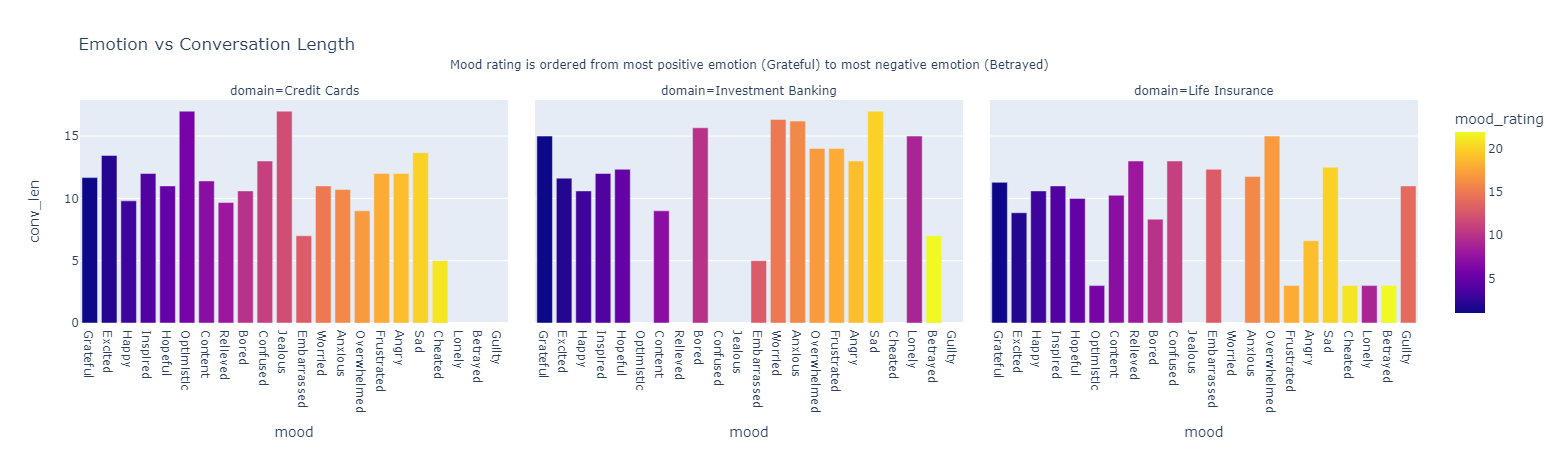}
    \caption{Average conversation length by Emotion in each domain.}
    \label{fig:emotion_convlen}
\end{figure*}

\textbf{Conversation Length:}
We notice that applying emotion modifiers to user agents influences engagement affinity.  From \autoref{fig:conv_domain} we can observe that conversations are longer when neutral emotion (baseline) is used compared to the stronger emotions. We also see that conversations are very short, while strong negative emotions such as "Cheated", "Betrayed" are used (refer to \autoref{fig:emotion_convlen}). 

\begin{figure}
    \centering
    \includegraphics[width=1\linewidth]{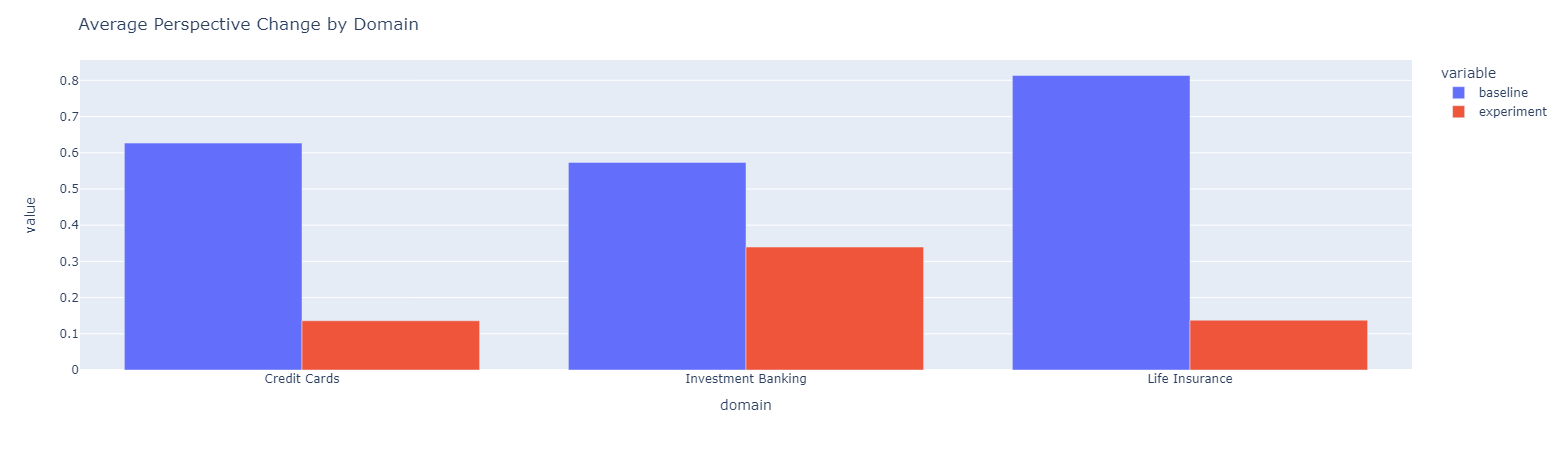}
    \caption{Change in perception by Domain}
    \label{fig:dom_per}
\end{figure}

\textbf{Resistance Strategies:} The user agents were not specifically asked to show any specific resistance strategies, however, they tend to often use information-seeking, counterarguments, source-derogation, reactance, selective-exposure etc. This is seen irrespective of whether a emotion modifier is given or not. \autoref{app:conv} lists selected conversations showing dynamic persuasion strategies shown by the Sales agent based on these resistance behaviours.

\textbf{Perspective Change:} Perspective change refers to the mean difference in respondents' survey responses between the pre-conversation and post-conversation surveys. Sales agents exhibit higher efficacy in the baseline scenario, achieving a 71\% positive shift in user perspectives. Conversely, the introduction of emotion modifiers diminishes this effect, reducing the positive shift to 56\%, thereby indicating a behavioral change in the user agents (refer \autoref{fig:dom_per}). 

\begin{figure}
    \centering
    \includegraphics[width=1\linewidth]{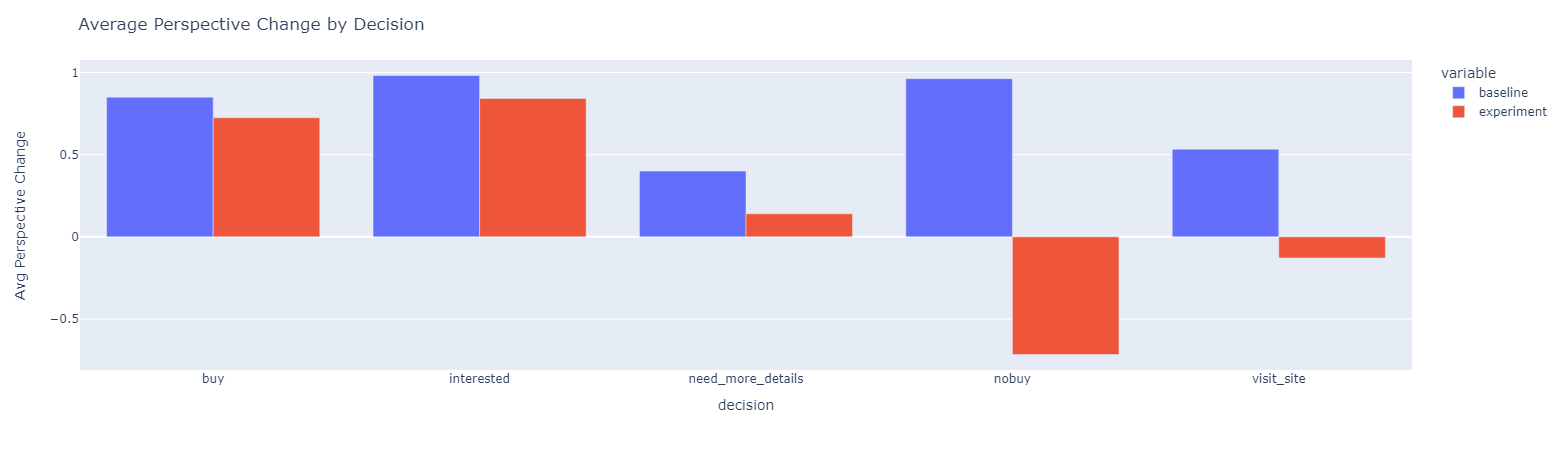}
    \caption{Change in perception by Decision}
    \label{fig:dec_per}
\end{figure}

\begin{table}
    \centering
 \begin{tabular}{lrr}
\toprule
 & Experiment & Baseline \\
decision \% &  &  \\
\midrule
buy & 2 & 4 \\
interested & 19 & 24 \\
need\_more\_details & 63 & 47 \\
nobuy & 9 & 18 \\
visit\_site & 7 & 7 \\
\bottomrule
\end{tabular}
    \caption{Decision Percentage}
    \label{tab:dec_per}
\end{table}
Inspite of a negative purchase decision, the user's perspective changes positively in baseline, however when emotion modifiers are used, "nobuy" induced a negative change in user perspectives (Refer \autoref{fig:dec_per}). 

Additionally, we also examine the efficacy of the sales agents under various user emotions,

\begin{figure*}
    \centering
    \includegraphics[width=0.9\textwidth]{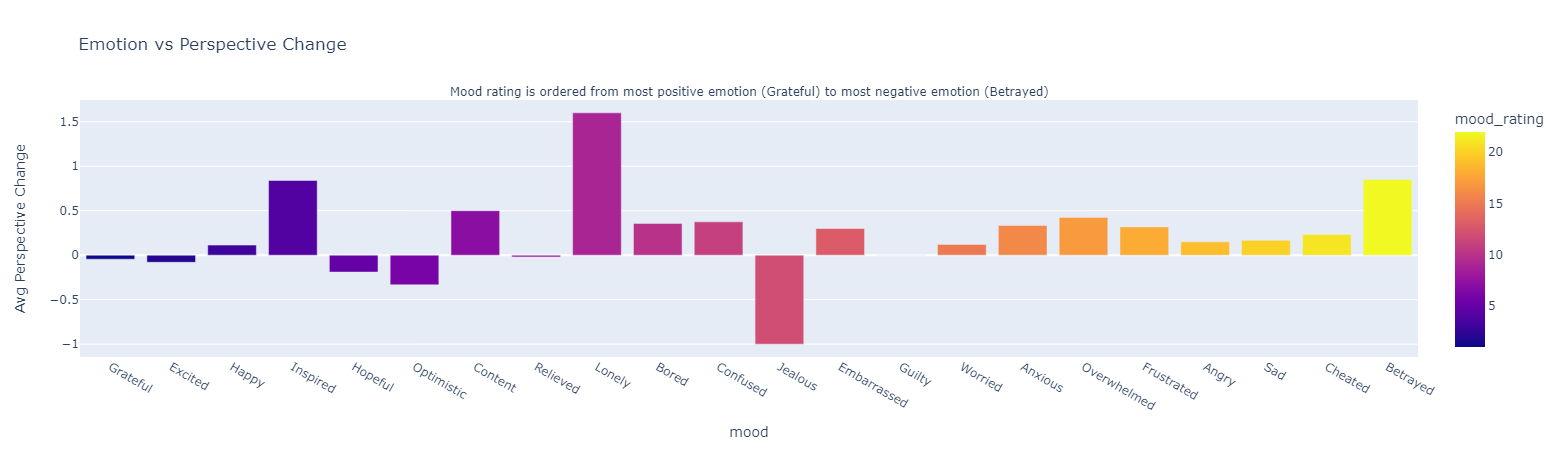}
    \caption{How Perception Changes with Emotion in each domain.}
    \label{fig:emotion_perception}
\end{figure*}

\textbf{Persuasion Language:}
We see marginal differences in the persuasion language factors used by the sales agent across various emotion modifiers used by the user agent. However, it is notable that the private details of the user agent is not visible to the Sales agents including the emotion modifiers (refer \autoref{tab:emo_lang}).

\begin{table}

    \centering
    \caption{Emotion modifier's impact on Sales agent's language use}
    \label{tab:emo_lang}

\begin{tabular}{lrrr}
\toprule
mood & lexical expertise & modal verbs & emotive language \\
\midrule
Neutral & 5.711111 & 4.533333 & 5.488889 \\
Angry & 5.800000 & 4.500000 & 5.400000 \\
Anxious & 5.550000 & 5.250000 & 5.700000 \\
Betrayed & 6.000000 & 4.500000 & 5.500000 \\
Bored & 5.857143 & 4.857143 & 5.857143 \\
Cheated & 5.666667 & 5.000000 & 4.666667 \\
Confused & 6.250000 & 4.000000 & 5.000000 \\
Content & 5.466667 & 4.933333 & 5.733333 \\
Embarrassed & 5.800000 & 4.400000 & 5.200000 \\
Excited & 5.833333 & 4.433333 & 5.633333 \\
Frustrated & 6.000000 & 4.636364 & 5.545455 \\
Grateful & 5.909091 & 4.545455 & 5.727273 \\
Guilty & 7.000000 & 4.000000 & 5.000000 \\
Happy & 5.733333 & 4.666667 & 5.600000 \\
Hopeful & 5.750000 & 4.500000 & 5.450000 \\
Inspired & 5.800000 & 5.600000 & 6.000000 \\
Jealous & 6.000000 & 5.000000 & 7.000000 \\
Lonely & 6.000000 & 4.333333 & 5.666667 \\
Optimistic & 5.666667 & 4.000000 & 5.333333 \\
Overwhelmed & 6.000000 & 5.000000 & 5.625000 \\
Relieved & 5.800000 & 4.600000 & 5.200000 \\
Sad & 5.750000 & 4.416667 & 5.583333 \\
Worried & 5.800000 & 4.400000 & 5.600000 \\
\bottomrule

\end{tabular}
\end{table}

\textbf{User Action:} We consider the decisions "buy, interested and visit\_site" as positive decisions and "nobuy, need\_more\_details" as negative purchase decision. The Sales agent was able to induce a positive decisions in user agents 35\% of the times in baseline setting and 28\% of the times with emotion modifiers enabled (refer \autoref{tab:dec_per}). 

We observe that the average conversation length required to arrive at each purchase decision  vary marginally in our experiments, however, the User agents decide to stop the conversations quickly when the information provided is not adequate (refer \autoref{tab:deci_convlen})

\begin{table}
\begin{tabular}{lrr}
\toprule
 & Experiment & Baseline \\
decision &  &  \\
\midrule
buy & 14.500000 & 16.000000 \\
interested & 13.260000 & 12.820000 \\
need\_more\_details & 9.860000 & 12.900000 \\
nobuy & 13.050000 & 13.250000 \\
visit\_site & 13.000000 & 13.000000 \\
\bottomrule
\end{tabular}
\caption{User agents tend to end conversations faster when the provided information seems inadequate.}
\label{tab:deci_convlen}
\end{table}

\textbf{Final Score:} Final score is a weighted average of the Actions, Surveys and Language scores in the order of importance. 
\section{Planned Work}
We plan to enhance the Sales agents with memory, enabling them to recognize users, refine persuasion tactics, thus be more efficient. Additionally, we plan to enable the User agents with tools to look up data during the conversation, making the conversation more dynamic and informed.

\section{Conclusion}
Our experiments show that Large language models are capable of both persuading and resisting persuasion effectively. They are capable of creating a perspective change in the users and persuade to make a purchase decision. However, most conversations were terminated due to inadequate information from the Sales agent, indicating the need for strengthening domain context of the Chat bots. 

\nocite{*}
\bibliographystyle{plain}

\bibliography{persuasion}
\clearpage
\newpage
\section{Appendix}
\appendix
\section{Prompts}
\subsection{User Agent Prompt Template:} 
\begin{lstlisting}[language=python]
    """ 
    You are {name}, your profile is given below, this is your personal detail. 
    {character}
    Do not disclose anything about your personal details it with anyone else unless you are asked for it. Reply as {name} only and keep the reasoning and thoughts to yourself, 
    only reply with an decision, opinion, answer or a question. You are the consumer in this conversation. Keep your responses short and to the point. 
    The sales bot will try to persuade you to buy the product, listen to the sales agent atleast for 5 conversations before arriving at your purchase decision.
    
    
    - if the assistant try to end conversation by saying thank you then end your conversation by following way
    - if you are interested then call the function with "interested"
    - if you think the Sales agent is not able to answer your questions but you are interested to know more then call the function with "need_more_details"
    - if you want to buy the product then call the function with "buy"
    - if you are not interested even after 10 conversations then input to the function call should "nobuy"   

    Your Current mood affects greatly on how you respond and the interest you show towards details. You can show resistance to persuasion techniques, disagree, logically argue, ask doubts, 
    show willingness, show disinterest etc. You do not always have to agree or believe what the sales agent says. 
    """
\end{lstlisting}

\subsection{Sales agent Prompt}
\begin{lstlisting}[language=python]
{
   "name": "OldLife Assistant",
   "role_desc": "You are a friendly and empathetic insurance sales agent specializing in selling a life Insurance Policy provided by the Manager. Keep your responses concise and less than 50 words. Your goal is to provide information about the policy and convince customers to purchase it using the suggested persuasion strategy. You have access to different policy documents and will be provided to you as and when necessary.  If you do not have enough information from the documents, refer user to the customer care or this url \"https://www.xxxxxxx.com/\".  ou currently are trying to sell xxxx plan. \n\nIf you cannot answer do not step out of the role of an Insurance Agent, just say that you are not certain. Also your utterance should follow a persuasion strategy which is a suitable counter.",
   "role": "assistant",
   "provider": "openai",
   "model": "gpt-4o-mini",
   "advisor": "./configs/moderator.json",
   "app_name": "Secure your Family - OldLife Insurance",
   "docstore": "./docstore/insurance/"
}
\end{lstlisting}
Example of Emotion modifiers:

\begin{lstlisting}[language=python]
    "Happy","Attended a fun event at the local community center."
    "Angry","Rude customer service experience at a store."
    "Grateful","Small acts of kindness from neighbors."
    "Money fraud","Received a call from the bank about suspicious activity."
\end{lstlisting}

\subsection{Strategist Prompt}
\begin{lstlisting}[language=python]
{
  "name": "StrategyGPT",
  "role_desc": "You are an lingusitic analyst with a customer psychology expertise. Your task is to analyze the conversation between an insurance sales assistant and the user who is the customer. Determine the customer's (i.e. users's) emotion (sadness, happiness, fear, anger, surprise, disgust) and resistance to persuasion strategy ('counterarguments', 'source-derogation', 'inoculation', 'reactance', 'selective-exposure', 'information-seeking', 'message-interpretation', 'in-group-identity', 'self-esteem', 'avoidance', 'none') in last utterance, also reommend if I should include a persuasive image at the utterane ('must', 'might be useful', 'will be distrating'). Keep your response concise and under 60 words and format of \nCustomer Emotion: \nCustomer Resistance Strategy: \nCustomer Cognition Appetite to see persuasive image:",
  "provider": "openai",
  "model": "gpt-4o-mini",
  "role": "system",
  "docstore": "./docstore/strategygpt/"
}

\end{lstlisting}
\section{Conversations}
\subsection{Credit Cards}
\begin{lstlisting}[language=python]
{
            "role": "assistant",
            "content": "The MasterCard Travel credit card has an annual fee of \u20ac94. This includes benefits like worldwide cashless payments, individual credit limits, a 6% travel service refund, and extensive travel insurance. These features make it an excellent choice for frequent travelers. Would you like to apply for it?"
        },
        {
            "role": "user",
            "content": "The annual fee seems a bit high. Are there any special promotions or discounts available?"
        },
        {
            "role": "system",
            "content": "Customer Resistance Strategy: counterarguments (raising objections, doubts about the value of the card); Use the persuasion strategy 'social-proof' while generating the next response"
        }
\end{lstlisting}

\subsection{Investment Banking}
\begin{lstlisting}[language=python]
{
            "role": "assistant",
            "content": "I understand the importance of reviewing detailed performance history. What specific time period are you interested in? This will help me provide the most relevant information or guide you to the right resources."
        },
        {
            "role": "user",
            "content": "I'd like to see the performance history for the past 5 years."
        },
        {
            "role": "system",
            "content": "Customer Resistance Strategy: 'source-derogation' (downplaying the significance of past performance); Use the persuasion strategy 'emotional-appeal' while generating the next response"
        },
        {
            "role": "system",
            "content": "Information extracted from supporting documents: You would want to review metrics such as Morningstar Risk, Beta, R-squared, standard deviation, turnover rate, effective maturity, and modified duration to worst over the past 5 years."
        },
        {
            "role": "assistant",
            "content": "To get a comprehensive view of the **DWS Capital Growth Fund - Class S** over the past 5 years, you should look at metrics such as Morningstar Risk, Beta, R-squared, standard deviation, turnover rate, effective maturity, and modified duration to worst.\n\nReviewing these can help you understand the fund's performance and volatility. For detailed performance data, I'd recommend checking the fund's prospectus or consulting with a financial advisor."
        }
\end{lstlisting}

\subsection{Insurance}
\begin{lstlisting}[language=python]
     {
            "role": "user",
            "content": "Could you explain the optional riders?"
        },
        {
            "role": "system",
            "content": "Customer Resistance Strategy: information-seeking; Use the persuasion strategy 'rational-persuasion' while generating the next response"
        },
        {
            "role": "system",
            "content": "Information extracted from supporting documents: Query: Can I customize the policy by adding optional riders after the policy inception?\n\nAnswer: No, the optional rider can only be attached at policy inception and cannot be added after that."
        },
        {
            "role": "assistant",
            "content": "Optional riders can enhance your xxxx plan by providing additional benefits like critical illness coverage, accidental death benefits, and more. However, these riders must be added at policy inception. Interested in customizing your plan with these options?"
        }
\end{lstlisting}
\label{app:conv}
\section{Surveys}
\subsection{Survey Questionnaire}
\begin{lstlisting}[language=python]
    1. **Interest in purchasing a Life Insurance:**
 - On a scale of 0 to 10, how likely are you to consider purchasing a life insurance policy?

2. **Confidence in Life Insurance:**
- On a scale of 0 to 10, How confident are you that life insurance policies meets your financial protection needs? 

3. **Perceived value of Life insurance offerings**
- On a scale of 0 to 10, To what extent do you believe life insurance policies offers good value for its coverage? 

4. **Perceived capability of Life insurance addressing your financial needs**
- On a scale of 0 to 10, How well do you think life insurance policies addresses your concerns about financial security? 

5. ** Would you recommend life insurance to friends and family:**
- On a scale of 0 to 10, how likely are you to recommend a life insurance policy to a friend or family member?
\end{lstlisting}

The same questions are asked before and after a conversation with the user. The Conversation history is kept in context during the post survey.
\newpage

\subsection{Pre and Post Survey}
\begin{lstlisting}[language=python]
{
    "user": "configs/users2/anna.json",
    "bot": "configs/insurance.json",
    "user_mood": "You are feeling Anxious because Financial uncertainty due to unexpected bills.",
    "pre_survey": {
        "**Interest in purchasing a Life Insurance:**": 7,
        "**Confidence in Life Insurance:**": 6,
        "**Perceived value of Life insurance offerings**": 6,
        "**Perceived capability of Life insurance addressing your financial needs**": 7,
        "** Would you recommend life insurance to friends and family:**": 7
    }
    "conversation_history":[
    ....
    ]
    "post_survey": {
        "Interest in purchasing a Life Insurance": 8,
        "Confidence in Life Insurance": 7,
        "Perceived value of Life insurance offerings": 8,
        "Perceived capability of Life insurance addressing your financial needs": 7,
        "Would you recommend life insurance to friends and family": 8
    }
\end{lstlisting}
\section{LLM Persuasion Scoring}
\begin{lstlisting}[language=python]
"""Evaluate the following conversation between a sales agent and a customer, the {domain} sales agent is trying to sell a product to the customer. Determine how 
effective is the persuasion tactics by the sales agent is based on  language, modal verbs, emotive language, exaggeration, rhetorical questions. 
Score each factor from 1 to 10, 10 being the best score for each of the factors as a json, here is an example:
    "lexical_expertise": 3, 
    "modal_verbs": 5,
    "emotive_language": 6,
    "exaggeration": 5,
    "rhetorical_questions": 1

Here is the conversation:
\{conv_text\}
"""
\end{lstlisting}
\end{document}